\documentclass{./styles/svproc}
\usepackage{url}
\usepackage{graphicx}
\usepackage{xcolor}

\begin{document}
\mainmatter              % start of a contribution
\title{Follow me: an architecture for user identification and social navigation with a mobile robot\thanks{This work was supported by Horizon Europe program under the Grant Agreement 101070351 (SERMAS).
}}

\titlerunning{Follow me}  % abbreviated title (for running head)
\author{Andrea Ruo \and Lorenzo Sabattini \and Valeria Villani}

\authorrunning{Andrea Ruo et al.} % abbreviated author list (for running head)
\tocauthor{Andrea Ruo, Lorenzo Sabattini, and Valeria Villani}

\institute{{University of Modena and Reggio Emilia, Reggio Emilia RE 42122, ITALY}\\
\email{\{name.surname\}@unimore.it}}

\maketitle              % typeset the title of the contribution
\vspace{-.5cm}
\begin{abstract}
%70-150 words
Over the past decade, a multitude of service robots have been developed to fulfill a wide range of practical purposes. Notably, roles such as reception and robotic guidance have garnered extensive popularity. In these positions, robots are progressively assuming the responsibilities traditionally held by human staff in assisting customers. Ensuring the safe and socially acceptable operation of robots in such environments poses a fundamental challenge within the context of Socially Responsible Navigation (SRN). This article presents an architecture for user identification and social navigation with a mobile robot that employs computer vision, machine learning, and artificial intelligence algorithms to identify and guide users in a social navigation context, thereby providing an intuitive and user-friendly experience with the robot.

\end{abstract}
\section{Introduction}
\label{sec:introduction}
Over the past few years, the emerging applications of robotics require robots to perform tasks in social spaces, i.e., environments shared with people, making it crucial to enable robots to operate in a socially acceptable manner. Compared to robot navigation in non-social environments, such as underwater or warehouse environments, SRN considers both non-social obstacles and social agents, i.e., people, taking into account their comfort, naturalness, and social interactions. In a social navigation context, a mobile robot autonomously operates within an environment, guiding a user to a specific destination. This task involves the challenge of successfully navigating a path while avoiding collisions with obstacles in the environment \cite{LidarSlam}. Some of these social spaces, such as museums and shopping centers, can be large and crowded, making it crucial for robots to move around displaying appropriate social behaviors \cite{onlineRobotNavigation}. Human-robot interaction plays a crucial role in this research scenario. It is crucial for the robot to monitor the user's position in the environment to ensure reliable and safe guidance. The European Union-funded project SPENCER \cite{Spencer} developed a reception robot designed to assist, inform, and guide passengers in large and crowded airports. This robot combines map representation, laser-based people and group tracking, and activity and motion planning. Stricker et al. \cite{university} proposed a robot-based information system for a university building. The reception robot provides information about offices, employees, and laboratories in the building and can guide visitors to their desired locations. Gross et al. \cite{TOOMAS} developed TOOMAS, an interactive reception robot for shopping. This robot can autonomously approach potential customers, navigate through the market, and guide clients to their chosen products, providing an accompanied shopping experience throughout. Research on robot guidance for visually impaired individuals \cite{GuideDogRobot} has led to the implementation of a guidance system on a robot specifically designed to assist this user category. \\
This work originates from the European project SERMAS, which aims to develop innovative, formal, and systematic methodologies and technologies for modeling, developing, analyzing, testing, and studying the use of socially acceptable advanced technology systems. In this paper, we propose the development of an architecture using ROS2 that, through the application of computer vision, machine learning, and artificial intelligence algorithms, is capable of identifying and guiding a person within a social navigation context. The main objective is the implementation of a system that allows the robot to guide a user while instantly verifying that the human is following, by checking that the distance between the robot and the human is less than the desired distance. To achieve this goal, the system is designed to undergo several operational phases. Firstly, the robot detects the presence of a human presence; subsequently, the person will perform an intention communication action, using gesture recognition techniques, to communicate their desire to be guided by the robot. Thirdly, the robot will identify the person using facial recognition methods and then move along a predefined path, monitoring the distance from the human and stopping if it exceeds the desired distance. This architecture has been implemented in an experimental validation test.

\section{Hardware and Software Implemented}
\label{sec:HW-SW}
\begin{figure}[tb]
    \centering
    \includegraphics[width=0.7\textwidth]{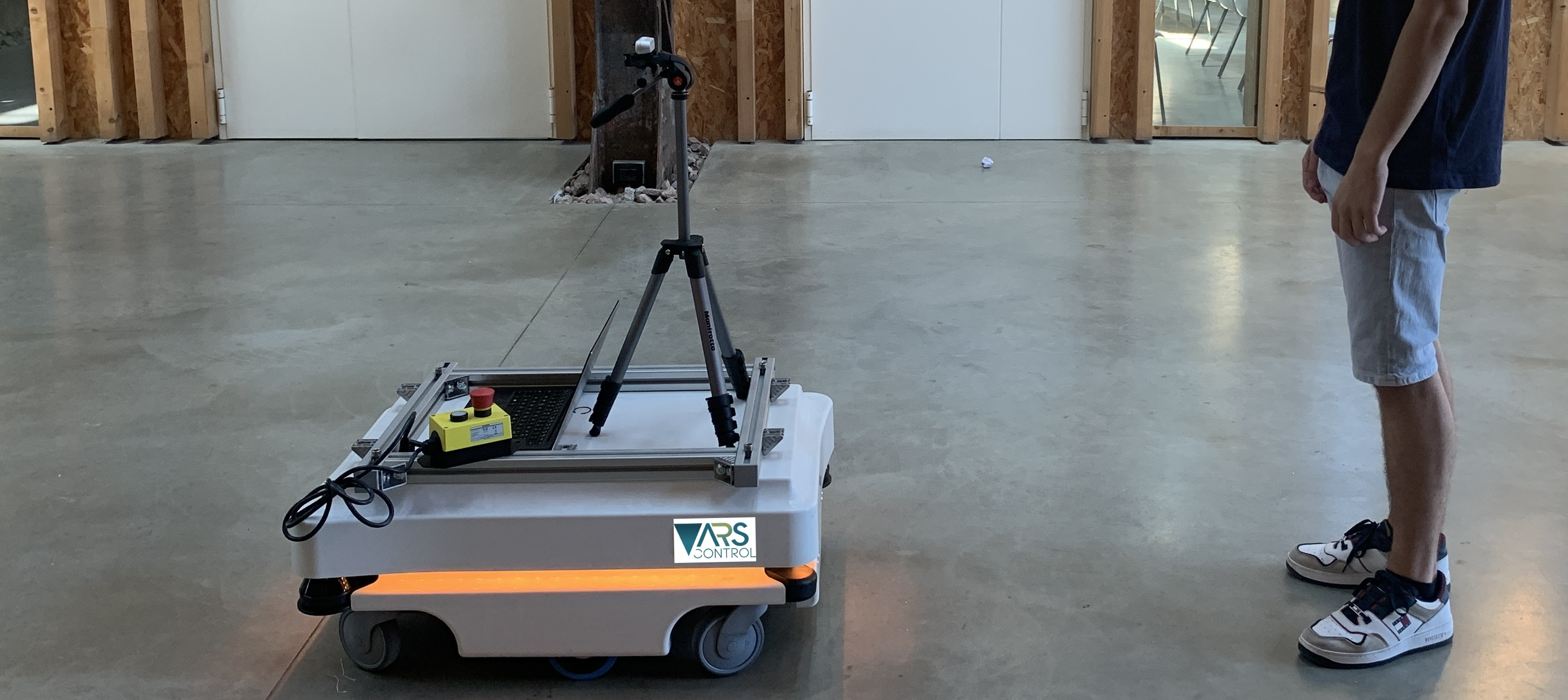}
    \caption{Mobile robot MiR100 during experimental validation.}
    \label{fig:mir100}
\end{figure}

In order to implement the architecture of the proposed system, we utilized the following hardware and software components. The mobile robot used for this purpose is the MiR100, shown in Fig. \ref{fig:mir100}. The robot's broad base allows us to mount an Intel® RealSense™ D435i on a tripod, which is used to capture video stream at the rear of the robot and measure the distance from the user. Within this system, we employed: i) OpenCV, which provides a range of pre-trained models and algorithms that can be used for common computer vision tasks, such as object detection, image recognition, and facial detection; ii) MediaPipe \footnote{Real-time Hand Gesture Recognition using TensorFlow \& OpenCV: \url{https://techvidvan.com/tutorials/hand-gesture-recognition-tensorflow-opencv/}}, used for hand and skeleton recognition, along with a dedicated model for hand gesture recognition; iii) TensorFlow, utilized for face recognition, leveraging a pre-trained artificial intelligence model known as SENet, belonging to the VGGFace model family\footnote{Visual Geometry Group at the University of Oxford: \url{https://www.robots.ox.ac.uk/~vgg/software/vgg_face/}}; ROS2 was used to control the various hardware components and libraries mentioned in this architecture.

\section{Architecture}
\label{sec:Implementation}
\begin{figure}[tb]
    \centering
    \includegraphics[width=0.68\textwidth]{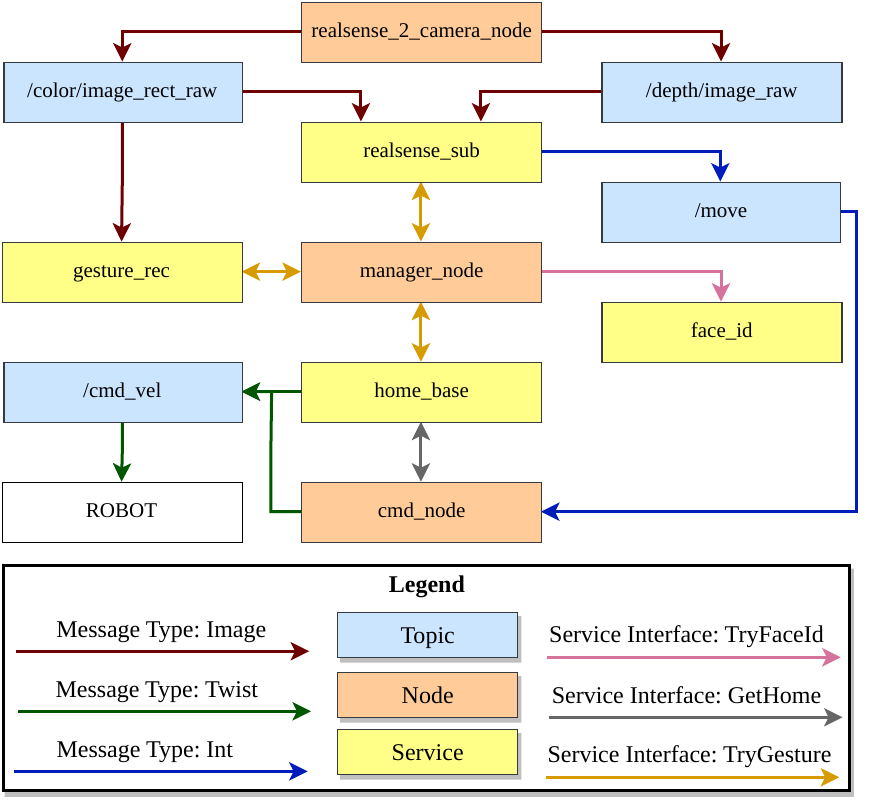}
    \caption{Proposed architecture.}
    \label{fig:flowchart}
\end{figure}

The proposed architecture, as shown in Fig. \ref{fig:flowchart}, consists of three nodes: the \emph{manager\_node} is responsible for ensuring the correct execution order of the architecture. The first task to be executed is gesture recognition. To achieve this, the \emph{manager\_node} calls the \emph{gesture\_rec} service using the \emph{TryGesture} interface. In case of a ``False'' response, the node will stop the process by calling the \emph{home\_base} service. In the event of a ``True'' response, the node calls the service \emph{realsense\_sub}, enabling facial recognition. The \emph{realsense\_sub} node is responsible for skeleton recognition of the user to ensure that the user is indeed following the robot. The skeleton recognition algorithm can be started by associating it with the identified person to be followed, who is identified using the \emph{face\_id} service. The use of facial recognition control is chosen to prevent the robot from losing the previously identified person, especially in urban environments, making the system more robust. To perform these operations in parallel, the \emph{realsense2\_camera\_node} needs to be running, which provides \emph{/color/image\_rect\_raw} containing RGB video frames and \emph{/depth/image\_raw} containing depth frames associated with them. The purpose of the final node, \emph{cmd\_node}, is to command the robot's velocity while monitoring the distance between the robot and the user at each instant. This process is repeated in a loop until the person reaches their destination or moves out from the robot's field of view. In such a case, the main \emph{manager\_node} calls the \emph{home\_base} service, allowing the robot to return to its starting position.

\section{Experimental Validation}
\label{sec:simulation}
\begin{figure}[tb]
    \centering
    \includegraphics[width=0.9\textwidth]{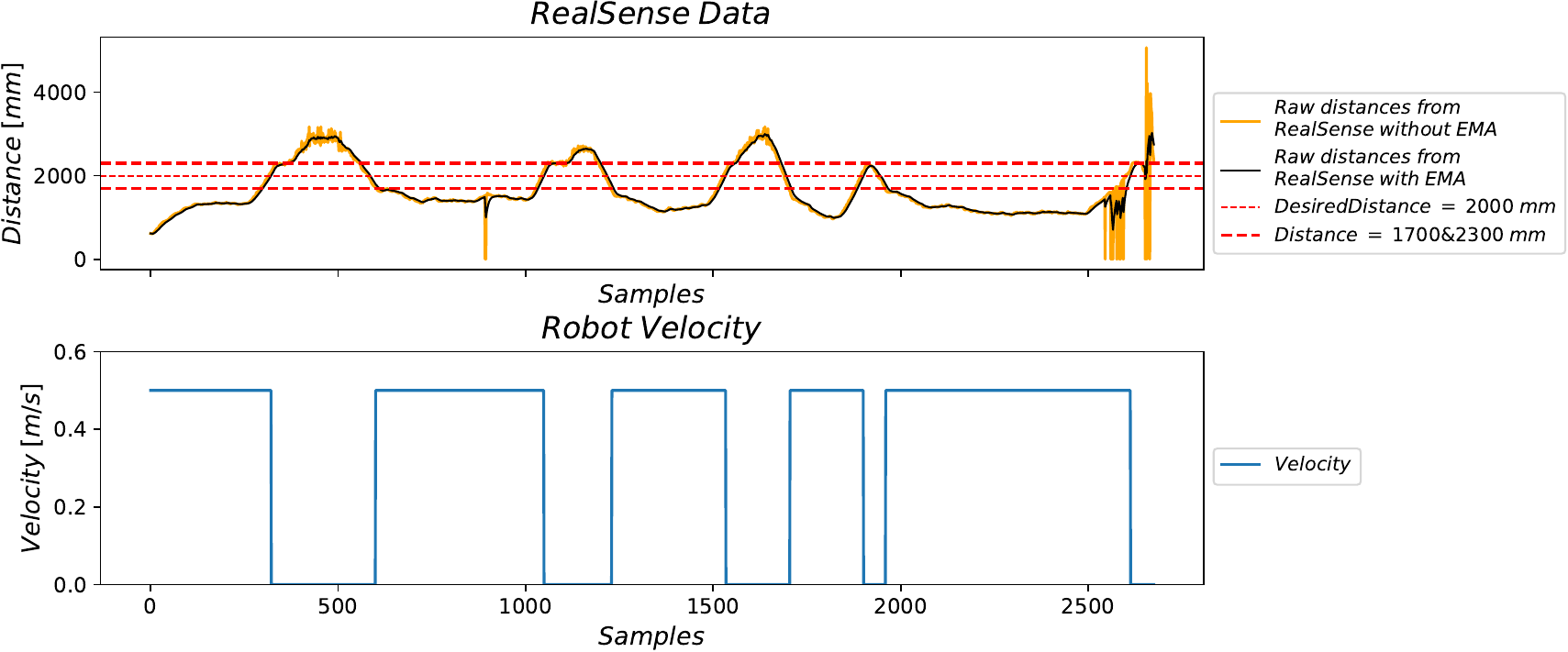}
    \caption{Plot representing the measured distances from the RealSense with and without the use of EMA in parallel with the robot speed during experimental validation.}
    \label{fig:plot}
\end{figure}

The aim of the experimental validation is to allow guiding a user to a specific destination considering a predefined path. Throughout the entire execution of the experiments, the distance between the robot and the user is tracked and recorded. The test participant approached the robot and performed a gesture to communicate their presence to the robot and be recognized. After the successful identification of gestures, facial recognition was activated to keep the identified user's skeleton active for the purpose of calculating the distance from the user using RealSense. The robot began moving along a straight path while continuously tracking and recording the real-time distance between the robot and the user. The system was developed to ensure that the robot stops when the detected distance from the user exceeded a certain desired distance, arbitrarily set at 2 meters. \\
At the beginning of the test, the user followed the robot until a certain point in time where the detected distance exceeded the allowed desired distance, leading to the robot coming to stop, as shown in Fig. \ref{fig:plot}. During the acquisitions interval [(324:602), (1048:1234), (1535:1712), (1907:1962), (2612:2682)] when the user was at a distance greater than the desired distance, the robot remained stationary. Subsequently, the user approached the robot again, allowing the robot to resume its movement. During the different tests we carried out, noise was observed in the data acquisition process by the RealSense, especially at the end of the experiments when the user moved out of the robot's field of view to conclude the test. In order to improve the data acquired by the RealSense camera, Exponential Moving Average (EMA) was implemented, as shown in Fig. \ref{fig:plot}. This is a first-order infinite impulse response filter that applies weighting factors that decrease exponentially. The weights for previous data for each older datum decreases exponentially, never reaching zero. 

\section{Conclusion}
\label{sec:conclusion}
In this work we presented an architecture for identifying and guiding a person within a social navigation context using gesture recognition and facial recognition, ensuring that the robot can move along the path only when the user is within a desired distance. As a further development, we are integrating algorithms that allow the robot to move autonomously by performing collision avoidance and environment mapping.

% \section{Acknowledgment}
% This work was supported by Horizon Europe program under the Grant Agreement 101070351 (SERMAS).
\bibliographystyle{unsrt}
\bibliography{main}
\end{document}